\def\year{2019}\relax
\documentclass[letterpaper]{article} 
\usepackage{aaai19}  
\usepackage{times}  
\usepackage{helvet}  
\usepackage{courier}  
\usepackage{url}  
\usepackage{graphicx}  
\usepackage{amsmath}
\usepackage{amsthm} 
\usepackage{amssymb}   
\usepackage{algorithm}
\usepackage{algpseudocode}
\usepackage{subfig}

\newtheorem{theorem}{Theorem}

\newtheorem{assumption}{Assumption}

\newtheorem{definition}{Definition}
\newtheorem{remark}{Remark}

\def\A{\mathcal{A}}
\def\B{B}

\def\I{\mathbf{I}}
\def\N{\mathcal{N}}
                                
\def\V{\field{V}}                                
\def\X{\mathcal{X}}

\def\E{\mathbb{E}}
\def\V{\mathbb{V}}

\DeclareMathOperator*{\argmin}{argmin}
\DeclareMathOperator*{\argmax}{argmax}

\begin{document}
%

\title{Balanced Linear Contextual Bandits}

\author{
Maria Dimakopoulou\textsuperscript{1}, 
Zhengyuan Zhou\textsuperscript{2}, 
Susan Athey\textsuperscript{3}, 
Guido Imbens\textsuperscript{3}
\\
\textsuperscript{1}Department of Management Science \& Engineering, Stanford University \\
\textsuperscript{2}Department of Electrical Engineering, Stanford University \\
\textsuperscript{3}Graduate School of Business, Stanford University \\
\texttt{\{madima,zyzhou,athey,imbens\}@stanford.edu}}

\maketitle
\begin{abstract}
Contextual bandit algorithms are sensitive to the estimation method of the outcome model as well as the exploration method used, particularly in the presence of rich heterogeneity or complex outcome models, which can lead to difficult estimation problems along the path of learning.
We develop algorithms for contextual bandits with linear payoffs that integrate balancing methods from the causal inference literature in their estimation to make it less prone to problems of estimation bias.
We provide the first regret bound analyses for linear contextual bandits with balancing and show that our algorithms match the state of the art theoretical guarantees.
We demonstrate the strong practical advantage of balanced contextual bandits on a large number of supervised learning datasets and on a synthetic example that simulates model misspecification and prejudice in the initial training data.
\end{abstract}

\section{Introduction}
Contextual bandits seek to learn a personalized treatment assignment policy in the presence of treatment effects that vary with observed contextual features.
In such settings, there is a need to balance the exploration of actions for which there is limited knowledge in order to improve performance in the future against the exploitation of existing knowledge in order to attain better performance in the present (see \cite{bubeck2012regret} for a survey). 
Since large amounts of data can be required to learn how the benefits of alternative treatments vary with individual characteristics, contextual bandits can play an important role in making experimentation and learning more efficient.
Several successful contextual bandit designs have been proposed \cite{auer-linrel}, \cite{li-linucb}, \cite{agrawal-lints}, \cite{agarwal-ilovetoconbandits}, \cite{bastani2015online}. 
The existing literature has provided regret bounds (e.g., the general bounds of \cite{russo-vanroy}, the bounds of  \cite{rigollet-nonparamtheory}, \cite{perchet-nonparamtheory}, \cite{slivkins-nonparamtheory} in the case of non-parametric function of arm rewards), has demonstrated successful applications (e.g., news article recommendations \cite{li-linucb} or mobile health \cite{lei-mhealth}), and has proposed system designs to apply these algorithms in practice \cite{agarwal-debt}.

In the contextual setting, one does not expect to see many future observations with the same context as the current observation, and so the value of learning from pulling an arm for this context accrues when that observation is used to estimate the outcome from this arm for a different context in the future.
Therefore, the performance of contextual bandit algorithms can be sensitive to the estimation method of the outcome model or the exploration method used.  
In the initial phases of learning when samples are small, biases are likely to arise in estimating the outcome model using data from previous non-uniform assignments of contexts to arms.
The bias issue is aggravated in the case of a mismatch between the generative model and the functional form used for estimation of the outcome model, or similarly, when the heterogeneity in treatment effects is too complex to estimate well with small datasets.  In that case methods that proceed under the assumption that the functional form for the outcome model is correct may  be overly optimistic about the extent  of the learning so far, and emphasize exploitation over exploration.
Another case where biases can arise occurs when training observations from certain regions of the context space are scarce (e.g., prejudice in training data if a non-representative set of users arrives in initial batches of data). 
These problems are common in real-world settings, such as in survey experiments in the domain of social sciences or in applications to health, recommender systems, or education.  
For example, early adopters of an online course may have different characteristics than later adopters.  

Reweighting or balancing methods address model misspecification by making the estimation ``doubly-robust,'', robust against misspecification of the reward function, important here, and robust against the specification of the propensity score (not as important here because in the bandit setting we know the propensity score).  The term ``doubly-robust'' comes from the extensive literature on offline policy evaluation \cite{scharfstein1999adjusting}; it means in our case that when comparing two policies using historical data, we get consistent estimates of the average difference in outcomes for segments of the context whether we have either a well-specified model of rewards  or not, as long as we   have a good model of the arm assignment policy (i.e., accurate propensity scores). Because in a contextual bandit the learner controls the arm assignment policy conditional on the observed context,  we therefore  have access to accurate propensity scores even in small samples. So, even when the reward model is severely misspecified,  the learner can be used to fsobtain unbiased estimates of the reward function for each range of values of the context. 

We suggest the integration of balancing methods from the causal inference literature \cite{imbens-ci} in online contextual bandits.
We focus on the domain of linear online contextual bandits with provable guarantees, such as LinUCB \cite{li-linucb} and LinTS \cite{agrawal-lints}, and we propose two new algorithms, \textit{balanced linear UCB (BLUCB)} and \textit{balanced linear Thompson sampling (BLTS)}. 
BLTS and BLUCB build on LinTS and LinUCB respectively and extend them
in a way that makes them less prone to problems of bias. The balancing will lead to lower estimated precision in the reward functions, and thus will emphasize exploration longer than the conventional linear TS and UCB algorithms, leading to more robust estimates.

The balancing technique is well-known in machine learning, especially in domain adaptation and  studies in learning-theoretic frameworks \cite{huang-ml}, \cite{zadrozny-ml}, \cite{cortes-ml}.
There is a number of recent works which approach contextual bandits through the framework of causality \cite{bareinboim-bandits}, \cite{bareinboim-fusion}, \cite{forney-fusion}, \cite{lattimore-causalbandit}. 
There is also a significant body of research that leverages balancing for offline evaluation and learning of contextual bandit or reinforcement learning policies from logged data \cite{strehl2010learning}, \cite{dudik-offline-1}, \cite{li-offline-1}, \cite{dudik-offline-2}, \cite{li-offline-2}, \cite{swaminathan-offline}, \cite{jiang-offline}, \cite{thomas-offline}, \cite{athey-offline}, \cite{kallus-offline}, \cite{wang-offline}, \cite{deshpande-offline}, \cite{kallus2018policy}, \cite{zhou2018offline}. 
In the offline setting, the complexity of the historical assignment policy is taken as given, and thus the difficulty of the offline evaluation and learning of optimal policies is taken as given. 
Therefore, these results lie at the opposite end of the spectrum from our work, which focuses on the online setting.
Methods for reducing the bias due to adaptive data collection have also been studied for non-contextual multi-armed bandits \cite{villar-online}, \cite{nie-online}, but the nature of the estimation in contextual bandits is qualitatively different. 
Importance weighted regression in contextual bandits was first mentioned in \cite{agarwal-ilovetoconbandits}, but without a systematic motivation, analysis and evaluation. To our knowledge, our paper is the first work to integrate balancing in the online contextual bandit setting, to perform a large-scale evaluation of it against direct estimation method baselines with theoretical guarantees and to provide a theoretical characterization of balanced contextual bandits that match the regret bound of their direct method counterparts. The effect of importance weighted regression is also evaluated in \cite{bietti2018contextual}, but this is a successor to the extended version of our paper \cite{dimakopoulou2017estimation}.

We prove that the regret bound of BLTS is $\tilde{O}\left(d\sqrt{K T^{1+\epsilon} / \epsilon}\right)$ and that the regret bound on BLUCB is $\tilde{O}\left(\sqrt{TdK}\right)$
where $d$ is the number of features in the context, $K$ is the number of arms and $T$ is the horizon. 
Our regret bounds for BLTS and BLUCB match the existing state-of-the-art regret bounds for LinTS \cite{agrawal-lints} and LinUCB \cite{chu2011contextual} respectively. 
We provide extensive and convincing empirical evidence for the effectiveness of BLTS and BLUCB (in comparison to LinTS and LinUCB) by considering the problem of multiclass classification with bandit feedback. 
Specifically, we transform a $K$-class classification task into a $K$-armed contextual bandit \cite{dudik-offline-1} and we use 300 public benchmark datasets for our evaluation.
It is important to point out that, even though BLTS and LinTS share the same theoretical guarantee, BLTS outperforms LinTS empirically. Similarly, BLUCB has a strong empirical advantage over LinUCB.
In bandits, this phenomenon is not uncommon. For instance, it is well-known that even though the existing UCB bounds are often tighter than those of Thompson sampling, Thompson sampling performs better in practice than UCB \cite{chapelle-tsucb}.
We find that this is also the case for balanced linear contextual bandits, as in our evaluation BLTS has a strong empirical advantage over BLUCB. 
Overall, in this large-scale evaluation, BLTS outperforms LinUCB, BLUCB and LinTS.
In our empirical evaluation, we also consider a synthetic example  that simulates in a simple way two issues of bias that often arise in practice, training data with non-representative contexts and model misspecification, and find again that BLTS is the most effective in escaping these biases.

\section{Problem Formulation \& Algorithms} \label{designs}

\subsection{Contextual Bandit Setting}
In the stochastic contextual bandit setting, there is a finite set of arms, $a \in \A$, with cardinality $K$. 
At every time $t$, the environment produces $(x_t, r_t) \sim D$, where $x_t$ is a $d$-dimensional context vector $x_t$ and $r_t = \left(r_t(1), \dots, r_t(K)\right)$ is the reward associated with each arm in $\A$.
The contextual bandit chooses arm $a_t \in \A$ for context $x_t$ and observes the reward only for the chosen arm, $r_t(a_t)$.
The optimal assignment for context $x_t$ is $a_t^* = \argmax_a\left\{\E[r_t(a) | x_t=x]\right\}$. 
The expected cumulative regret over horizon $T$ is defined as $R(T) \triangleq \E\left[\sum_{t=1}^{T}\left(r_t(a^*_t) - r_t(a_t)\right)\right]$.
At each time $t = 1, \dots, T$, the contextual bandit assigns arm $a_t$ to context $x_t$ based on the history of observations up to that time, $(x_\tau, a_\tau, r_\tau(a_\tau))_{\tau=1}^{t-1}$.
The goal is to find the assignment rule that minimizes $R(T)$.

\subsection{Linear Contextual Bandits} \label{lcb}

\begin{algorithm}[t]
	\caption{Balanced Linear Thompson Sampling} 
	\label{alg:BLTS}
	\begin{algorithmic}[1]
		\State \textbf{Input:} Regularization parameter $\lambda > 0$, propensity score threshold $\gamma \in (0, 1)$, constant $\alpha$ (deafult is 1)
		\State Set $\hat{\theta}_a \leftarrow \textbf{null}, \B_a \leftarrow \textbf{null}, \forall a \in \mathcal{A}$
		\State Set $X_a \leftarrow$ empty matrix, $r_a \leftarrow$ empty vector $\forall a \in \mathcal{A}$
		\For {$t = 1, 2, \dots, T$}
		\If {$\exists a \in \A \text{ s.t. }\hat{\theta}_a = \textbf{null}$ or $\B_a = \textbf{null}$}
		\State Select $a \sim \text{Uniform}(\A)$
		\Else
		\State Draw $\tilde{\theta}_a$ from $\N\left(\hat\theta_a, \alpha^2 \V(\hat\theta_a) \right)$ for all $a \in \A$ 
		\State Select $a = \arg\max\limits_{a\in\A} x_t^\top \tilde{\theta}_a$
		\EndIf
		\State Observe reward $r_t(a)$.
		\State Set $W_a \leftarrow $ empty matrix
		\For {$\tau = 1, \dots, t$}
		\State Compute $p_a(x_\tau)$ and set $w = \frac{1}{\max(\gamma, p_a(x_\tau))}$
		\State $W_a \leftarrow \text{diag}(W_a, w)$
		\EndFor
		\State $X_a \leftarrow [X_a : x_t^\top]$
		\State $B_a \leftarrow X_a^\top W_a X_a + \lambda \I$
		\State $r_a \leftarrow [r_a : r_t(a)]$ 
		\State $\hat{\theta}_a \leftarrow B_a^{-1} X_a^\top W_a r_a$
		\State $\V(\hat\theta_a)\leftarrow  B_a^{-1}\left((r_a-X_a^\top\hat\theta_a)^\top W_a(r_a-X_a^\top\hat\theta_a)\right) $
		\EndFor
	\end{algorithmic}
\end{algorithm}

Linear contextual bandits rely on modeling and estimating the  reward distribution corresponding to each arm $a \in \A$ given context $x_t = x$. Specifically the expected reward is assumed to be a linear function of the context $x_t$ with some unknown coefficient vector $\theta_a$, $\E[r_t(a) | x_t = x] = x^\top \theta_a$, and the variance is typically assumed to be constant
$\V[r_t(a) | x_t = x] = \sigma^2_a$.
In the setting we are studying, there $K$ models to be estimated, as many as the arms in $\A$.
At every time $t$, this estimation of $\theta_a$ is done separately for each arm $a$ on the history of observations corresponding to this arm, $(X_a, r_a) = \{\left(x_t, r_t(a_t)\right) \text{, } t: a_t = a\}$.
Thompson Sampling  \cite{thompson-ts}, \cite{scott-ts}, \cite{agrawal-ts}, \cite{russo2018tutorial} 
and Upper Confidence Bounds (UCB) \cite{lai-ucb}, \cite{auer-ucb} are two different 
methods which are highly effective in dealing with the exploration vs. exploitation trade-off in multi-armed bandits.
LinTS \cite{agrawal-lints} and LinUCB \cite{li-linucb} are linear contextual bandit algorithms associated with Thompson sampling and UCB respectively.

At time $t$, LinTS and LinUCB apply ridge regression with regularization parameter $\lambda$ to the history of observations $(X_a, r_a)$ for each arm $a \in \A$, in order to obtain an estimate $\hat{\theta}_a$ and its variance $\V_a(\hat\theta_a)$.
For the new context $x_t$, $\hat{\theta}_a$ and its variance are used by LinTS and LinUCB to obtain the conditional mean $\hat{\mu}_a(x_t)=x_t^\top\hat\theta_a$  of the reward associated with each arm $a \in A$, and its  variance $\V(\hat{\mu}_a(x_t))=x^\top_t \V(\hat\theta_a) x_t $.
LinTS assumes that the expected reward $\mu_a(x_t)$ associated with arm $a$ conditional on the context $x_t$ is Gaussian $ \N\left(\hat{\mu}_a(x_t), \alpha^2\V(\hat{\mu}_a(x_t)) \right)$, where $\alpha$ is an appropriately chosen constant.
LinTS draws a sample $\tilde{\mu}_a(x_t)$ from the distribution of each arm $a \in \cal$ and context $x_t$ is then assigned to the arm with the highest sample, $a_t = \argmax_a \{\tilde{\mu}_{a}(x_t)\}$. 
LinUCB uses the estimate $\hat{\theta}_a$ and its variance to compute upper confidence bounds for the expected reward $\mu_a(x_t)$ of context $x_t$ associated with each arm $a \in \A$ and assigns the context to the arm with the highest upper confidence bound, $a_t = \argmax_a\left\{\hat{\mu}_a(x_t) + \alpha \sqrt{\V(\hat{\mu}_a(x_t))}\right\}$, where $\alpha$ is an appropriately chosen constant.

\subsection{Linear Contextual Bandits with Balancing} \label{blcb}

\begin{algorithm}[t]
\caption{Balanced Linear UCB} 
\label{alg:BLUCB}
\begin{algorithmic}[1]
\State \textbf{Input:} Regularization parameter $\lambda > 0$, propensity score threshold $\gamma \in (0, 1)$, constant $\alpha$.
\State Set $\hat{\theta}_a \leftarrow \textbf{null}, \B_a \leftarrow \textbf{null}, \forall a \in \mathcal{A}$
\State Set $X_a \leftarrow$ empty matrix, $r_a \leftarrow$ empty vector $\forall a \in \mathcal{A}$
\For {$t = 1, 2, \dots, T$}
\If {$\exists a \in \A \text{ s.t. }\hat{\theta}_a = \textbf{null}$ or $\B_a = \textbf{null}$}
\State Select $a \sim \text{Uniform}(\A)$
\Else
\State Select $a = \arg\max\limits_{a\in\A} \left(x_t^\top \hat{\theta}_a + \alpha \sqrt{x_t^\top  \V(\hat\theta_a) x_t}\right)$
\EndIf
\State Observe reward $r_t(a)$.
\State Set $W_a \leftarrow $ empty matrix
\For {$\tau = 1, \dots, t$}
\State Estimate $\hat{p}_a(x_\tau)$ and set $w = \frac{1}{\max(\gamma, \hat{p}_a(x_\tau))}$
\State $W_a \leftarrow \text{diag}(W_a, w)$
\EndFor
\State $X_a \leftarrow [X_a : x_t^\top]$
\State $B_a \leftarrow X_a^\top W_a X_a + \lambda \I$
\State $r_a \leftarrow [r_a : r_t(a)]$ 
\State $\hat{\theta}_a \leftarrow B_a^{-1} X_a^\top W_a r_a$
\State $\V(\hat\theta_a)\leftarrow  B_a^{-1}\left((r_a-X_a^\top\hat\theta_a)^\top W_a(r_a-X_a^\top\hat\theta_a)\right) $
\EndFor
\end{algorithmic}
\end{algorithm}

In this section, we show how to integrate balancing methods from the causal inference literature in linear contextual bandits, in order to make estimation less prone to bias issues.

Balanced linear Thompson sampling (BLTS) and balanced linear UCB (BLUCB) are online contextual bandit algorithms that perform balanced estimation of the model of all arms in order to obtain a Gaussian distribution and an upper confidence bound respectively for the reward associated with each arm conditional on the context.
We focus on the method of inverse propensity weighting \cite{imbens-ci}.
The idea is that at every time $t$, the linear contextual bandit weighs each observation $(x_\tau, a_\tau, r_\tau(a_\tau))$, $\tau = 1, \dots, t$ in the history up to time $t$ by the inverse probability of context $x_\tau$ being assigned to arm $a_\tau$. This probability is called propensity score and is denoted as $p_{a_\tau}(x_\tau)$. Then, for each arm $a \in \A$, the linear contextual bandit weighs each observation $(x_\tau, a, r_\tau(a))$ in the history of arm $a$ by $w_{a} = 1 / p_{a}(x_\tau)$ and uses weighted regression to obtain the estimate $\hat{\theta}^\text{BLTS}_a$ with variance $\V(\hat\theta^\text{BLTS}_a)$.
In BLTS (Algorithm \ref{alg:BLTS}), the propensity scores are known because Thompson sampling performs probability matching, i.e., it assigns a context to an arm with the probability that this arm is optimal.
Since computing the propensity scores involves high order integration, they can be approximated via Monte-Carlo simulation. 
Each iteration draws a sample from the posterior reward distribution of each arm $a$ conditional on $x$, where the posterior is the one that the algorithm considered at the end of a randomly selected prior time period. 
The propensity score $p_a(x_\tau)$ is the fraction of the Monte-Carlo iterations in which arm $a$ has the highest sampled reward, where the arrival time of context $x_\tau$ is treated as random.  
For every arm $a$, the history $(X_a, r_a, p_a)$ is used to obtain a balanced estimate $\hat{\theta}^\text{BLTS}_a$ of $\theta_a$ and its variance $\V(\hat\theta^\text{BLTS}_a)$ which produce a normally distributed estimate of $\tilde{\mu}_a \sim \N\left(x_t^\top \hat{\theta}^\text{BLTS}_a, \alpha^2 x_t^\top \V(\hat\theta^\text{BLTS}_a) x_t\right)$ of the reward of arm $a$ for context $x_t$, where $\alpha$ is a parameter of the algorithm.

In BLUCB (Algorithm \ref{alg:BLUCB}), the observations are weighed by the inverse of estimated propensity scores.
Note that UCB-based contextual bandits have deterministic assignment rules and conditional on the context the propensity score is either zero or one.
But with the standard assumption that the arrival of contexts is random, at every time period $t$ the estimated probability $\hat{p}_a(x_\tau)$ 
is obtained by the prediction of the trained multinomial logistic regression model on  $(x_\tau, a_\tau)_{\tau=1}^{t-1}$.
Subsequently, $(X_a, r_a, \hat{p}_a)$ is used to obtain a balanced estimate $\hat{\theta}^{\text{BLUCB}}_a$ of $\theta_a$ and its variance $\V(\hat\theta^\text{BLUCB}_a)$. These are used to construct the upper confidence bound, $x_t^\top \hat{\theta}_a + \alpha \sqrt{x_t^\top \V(\hat\theta^\text{BLUCB}_a) x_t}$, for the reward of arm $a$ for context $x_t$,  where $\alpha$ is a constant. (For some results, e.g., \cite{auer2002using}, $\alpha$ needs to be slowly increasing in $t$.)

Note that $\hat{\theta}^\text{BLTS}_a$, $\V(\hat\theta^\text{BLTS}_a)$ and $\hat{\theta}^\text{BLUCB}_a$, $\V(\hat\theta^\text{BLUCB}_a)$ can be computed in closed form or via the bootstrap.

Weighting the observations by the inverse propensity scores reduces bias, but even when the propensity scores are known it increases variance, particularly when they are small. Clipping the propensity scores \cite{crump2009dealing} with some threshold  $\gamma$, e.g. $0.1$ helps control the variance increase. This threshold $\gamma$ is an additional parameter to BLTS (Algorithm \ref{alg:BLTS}) and BLUCB (Algorithm \ref{alg:BLUCB}) compared to LinTS and LinUCB.
Finally, note that one could integrate in the contextual bandit estimation other covariate balancing methods, such as the method of approximate residual balancing \cite{athey-arb} or the method of \cite{kallus-offline}. 
For instance, with approximate residual balancing one would use as weights
$w_a = \argmin_w \left\{ (1- \zeta) \lVert w\rVert_2^2 + \zeta \lVert \bar{x} - \textbf{X}_a^T w \rVert_\infty^2\right\}$ s.t. $\sum_{t: a_t = a} w_t = 1 \text{ and } 0 \leq w_t \leq n_a^{-2/3}$
where $\zeta \in (0, 1)$ is a tuning parameter, $n_a =  \sum_{t = 1}^{T} \textbf{1} \{a_t = a\}$ and $\bar{x} = \frac{1}{T} \sum_{t = 1}^{T} x_t$ and then use $w_a$ to modify the parametric and non-parametric model estimation as outlined before. 

\section{Theoretical Guarantees for BLTS and BLUCB} \label{theory}
In this section, we establish theoretical guarantees of BLTS and BLUCB that are comparable to LinTS and LinUCB. We start with a few technical assumptions that are standard in the contextual bandits literature.

\begin{assumption}\label{assump:realizability}
\textbf{Linear Realizability:}
There exist parameters $\{\theta_a\}_{a\in \A}$ such that given any context $x$, $\mathbb{E}[r_t(a) | x] = x^\top \theta_a, \forall a \in \A, \forall t \ge 0$.
\end{assumption}

We use the standard (frequentist) regret criterion and standard assumptions on the regularity of the distributions.
\begin{definition}
The instantaneous regret at iteration $t$ is $x^\top_t \theta_{a_t^*} -  x^\top_t \theta_{a_t}$, 
where $a^*_t$ is the optimal arm at iteration $t$ and $a_t$ is the arm taken at iteration $t$.
The cumulative regret $R(T)$ with horizon $T$ is the defined as $R(T) = \sum_{t=1}^\top \left(x^\top_t \theta_{a_t^*} -  x^\top_t \theta_{a_t}\right)$.
\end{definition}

\begin{definition}
We denote the canonical filtration of the underlying contextual bandits problem by $\{\mathcal{F}_t\}_{t=1}^{\infty}$, where $\mathcal{F}_t = \sigma(\{x_s\}_{s=1}^t, \{a_s\}_{s=1}^t, \{r_s(a_s)\}_{s=1}^t, x_{t+1})$: the sigma algebra\footnote{All the random variables $x_t, a_t, r_t$ are defined on some common underlying probability space, which we do not write out explicitly here.} generated by all the random variables up to and including iteration $t$, plus $x_{t+1}$. In other words, $\mathcal{F}_t$ contains all the information that is available before making the decision for iteration $t+1$. 
\end{definition}

\begin{assumption}\label{assump:reg}
For each $a \in \A$ and every $t \ge 1$:
\begin{enumerate}
\item 
\textbf{Sub-Gaussian Noise:}
$r_t(a) - x_t^\top\theta_a$ is conditionally sub-Gaussian:
there exists some $L_a > 0$, such that
$\mathbb{E} [e^{s(r_t(a) - x_t^\top\theta_a)} \mid \mathcal{F}_{t-1}] \le \exp(\frac{s^2 L_a^2}{2}), \forall s, \forall x_t$.
\item
\textbf{Bounded Contexts and Parameters:}
The contexts $x_t$ and parameters $\theta_a$ are assumed to be bounded.
Consequently, without loss of generality, we can rescale them such that
$\|x_t\|_2 \le 1, \|\theta_a\|_2 \le 1, \forall a, t$.
\end{enumerate}

\end{assumption}

\begin{remark}
Note that we make no assumption of the underlying $\{x_t\}_{t=1}^{\infty}$ process:
the contexts $\{x_t\}_{t=1}^{\infty}$ need not to be fixed beforehand or come from some stationary process. Further, they can even be adapted to
$\sigma(\{x_s\}_{s=1}^t, \{a_s\}_{s=1}^t, \{r_s(a_s)\}_{s=1}^t)$, in which case they are called adversarial contexts in the literature as the contexts can be chosen by an adversary who chooses a context after observing the arms played and the corresponding rewards.
If $\{x_t\}_{t=1}^{\infty}$ is an IID process, then the problem is known as stochastic contextual bandits. From this viewpoint, adversarial contextual bandits are more general, but the regret bounds tend to be worse. Both are studied in the literature.
\end{remark}

\begin{theorem}\label{thm:main}
Under Assumption~\ref{assump:realizability} and Assumption~\ref{assump:reg}:
\begin{enumerate}
\item 
If BLTS is run with $\alpha = \sqrt{\frac{\log\frac{1}{\delta}}{\epsilon}}$ in Algorithm \ref{alg:BLTS},
then with probability at least $1 - \delta$, 
$R(T) = \tilde{O}\left(d\sqrt{\frac{KT^{1+\epsilon}}{\epsilon}}\right)$.
\item If BLUCB is run with $\alpha = \sqrt{\log \frac{TK}{\delta}}$ in Algorithm \ref{alg:BLUCB},
then with probability at least $1 - \delta$, 
$R(T) = \tilde{O}\left(\sqrt{TdK}\right)$.
\end{enumerate}
\end{theorem}
\noindent We refer the reader to Appendix A of the supplemental material of the extended version of this paper \cite{dimakopoulou2017estimation} for the regret bound proofs.

\begin{remark}
The above bound essentially matches the existing state-of-the art regret bounds for linear Thompson sampling with direct model estimation (e.g.~\cite{agrawal-lints}). Note that in~\cite{agrawal-lints},
an infinite number of arms is also allowed, but all arms share the same parameter $\theta$. The final regret bound is $\tilde{O}\left(d^2 \frac{\sqrt{T^{1+\epsilon}}}{\epsilon}\right)$. Note that even though no explicit dependence on $K$ is present in the regret bound (and hence our regret bound appears as a factor of $\sqrt{K}$ worse), this is to be expected, as we have $K$ parameters to estimate, one for each arm. Note that here we do not assume any structure on the $K$ arms; they are just $K$ stand-alone parameters, each of which needs to be independently estimated. 
Similarly, for BLUCB, our regret bound is $\tilde{O}\left(\sqrt{TdK}\right)$,
which is a factor of $\sqrt{K}$ worse than that of \cite{chu2011contextual},
which establishes a $\tilde{O}\left(\sqrt{Td}\right)$ regret bound.
Again, this is because a single true $\theta^*$ is assumed in \cite{chu2011contextual}, rather than $K$ arm-dependent parameters.

Of course, we also point out that our regret bounds are not tight, nor do they achieve state-of-the-art regret bounds in contextual bandits algorithms in general.
The lower bound $\Omega({\sqrt{dT}})$ is established in~\cite{chu2011contextual} for linear contextual bandits (again in the context of a single parameter $\theta$ for all $K$ arms).
In general, UCB based algorithms (\cite{auer-linrel,chu2011contextual,bubeck2012regret,abbasi2011improved}) tend to have better
(and sometimes near-optimal) theoretical regret bounds.
In particular, the state-of-the-art bound of $O(\sqrt{dT\log K})$ for linear contextual bandits is given 
in~\cite{bubeck2012regret} (optimal up to a $O(\log K)$ factor).
However, as mentioned in the introduction, Thompson sampling based algorithms tend to perform much better in practice (even though their regret bounds tend not to match UCB based algorithms, as is also the case here). Hence, our objective here is not to provide
state-of-the-art regret guarantees. Rather, we are motivated to design algorithms that have better empirical performance (compared to both the existing UCB style algorithms and Thompson sampling style algorithms), which also enjoy the baseline theoretical guarantee.

Finally, we give some quick intuition for the proof. For  BLTS, we first show that estimated means concentrate around true mean (i.e. $x_t^\top \hat{\theta}_a$ concentrates around $x_t^\top \theta_a$). Then, we establish that sampled means concentrate around the estimated means (i.e. $x_t^\top\tilde{\theta}_a$  concentrates around $x_t^\top\hat{\theta}_a$). These two steps together indicate that the sampled mean is close to the true mean. A further consequence of that is we can then bound the instantaneous regret (regret at each time step $t$) in terms of the sum of two standard deviations: one corresponds to the optimal arm at time $t$, the other corresponds to the actual selected arm at $t$. The rest of the proof then follows by giving tight characterizations of these two standard deviations. For BLUCB, the proof again utilizes the first concentration mentioned above: the estimated means concentrate around true mean (note that there is no sampled means in BLUCB). The rest of the proof adopts a similar structure as in~\cite{chu2011contextual}.
\end{remark}

\section{Computational Results} 

In this section, we present computational results that compare the performance of our balanced linear contextual bandits, BLTS and BLUCB, with the direct method linear contextual bandit algorithms that have theoretical guarantees, LinTS and LinUCB.
Our evaluation focuses on contextual bandits with linear realizability assumption and strong theoretical guarantees.
First, we present a simple synthetic example that simulates bias in the training data by under-representation or over-representation of certain regions of the context space and investigates the performance of the considered linear contextual bandits both when the outcome model of the arms matches the true reward generative process and when it does not match the true reward generative process.
Second, we conduct an experiment by leveraging 300 public, supervised cost-sensitive classification datasets to obtain contextual bandit problems, treating the features as the context, the labels as the actions and revealing only the reward for the chosen label.
We show that BLTS performs better than LinTS and that BLUCB performs better than LinUCB. 
The randomized assignment nature of Thompson sampling facilitates the estimation of the arms' outcomes models compared to UCB, and as a result LinTS outperforms LinUCB and BLTS outperforms BLUCB.
Overall, BLTS has the best performance.
In the supplemental material, we include experiments against the policy-based contextual bandit from \cite{agarwal-ilovetoconbandits} which is statistically optimal but it is also outperformed by BLTS.

\subsection{A Synthetic Example}\label{subsec:pedagogic}

This simulated example aims to reflect in a simple way two issues that often arise in practice.
The first issue is the presence of bias in the training data by under-representation or over-representation of certain regions. A personalized policy that is trained based on such data and is applied to the entire context space will result in biased decisions for certain contexts.  The second issue is the problem of mismatch between the true reward generative process and the functional form used for estimation of the outcome model of the arms, which is common in applications with complex generative models. Model misspecification aggravates the presence of bias in the learned policies.

We use this simple example to present in an intuitive manner why balancing and randomized assignment rule help with these issues, before moving on to a large-scale evaluation of the algorithms in real datasets in the next section.

\begin{figure}[!htb]
\centering
\includegraphics[width=0.5\columnwidth]{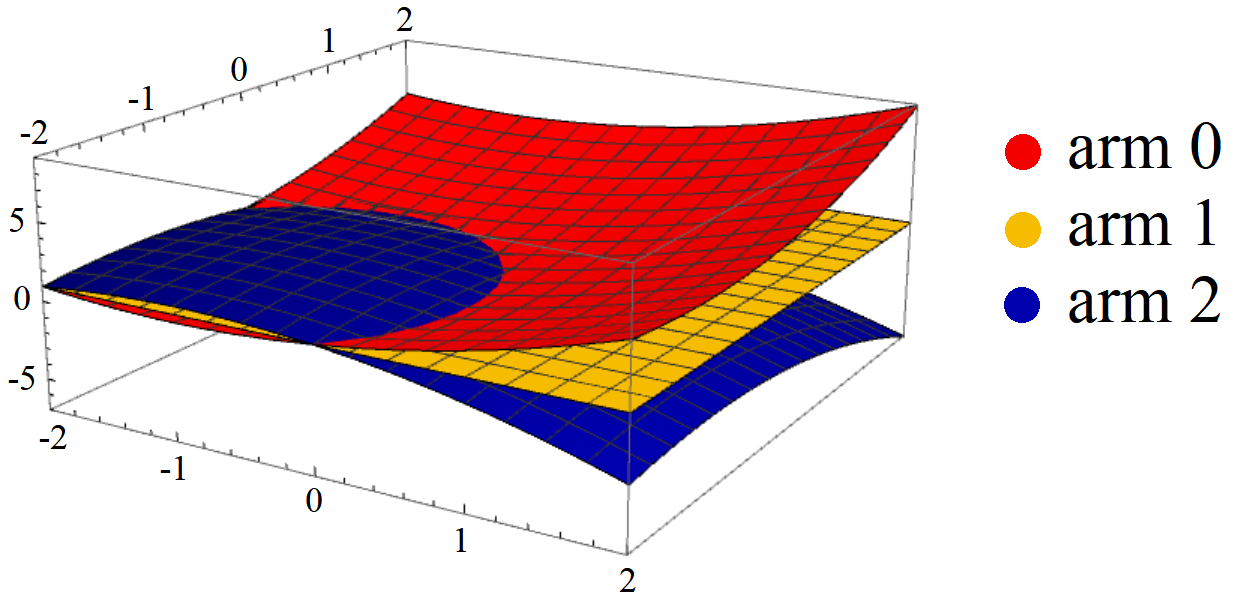}
\caption{Expectation of each arm's reward, $\E[r_t(0)] = 0.5 (x_{t,0} + 1)^2 + 0.5 (x_{t,1} + 1)^2$ (red), $\E[r_t(1)] = 1$ (yellow), $\E[r_t(2)] = 2 - 0.5 (x_{t,0} + 1)^2 - 0.5 (x_{t,1} + 1)^2$ (blue).}
\label{PotentialOutcomes}
\end{figure}

Consider a simulation design where there is a warm-start batch of training observations, but it consists of contexts focused on one region of the context space. 
There are three arms $\A = \{0, 1, 2\}$ and the contexts $x_t = (x_{t,0}, x_{t,1})$ are two-dimensional with $x_{t,j} \sim \mathcal{N}(0, 1), \enskip j \in \{0, 1\}$. 
The rewards corresponding to each arm $a \in \A$ are generated as follows; $r_t(0) = 0.5 (x_{t,0} + 1)^2 + 0.5 (x_{t,1} + 1)^2 + \epsilon_t$, $r_t(1) = 1 + \epsilon_t$, and $r_t(2) = 2 - 0.5 (x_{t,0} + 1)^2 - 0.5 (x_{t,1} + 1)^2 +\epsilon_t$, where $\epsilon_t \sim \mathcal{N}(0, \sigma^2)$, $\sigma^2 = 0.01$. 
The expected values of the three arms' rewards are shown in Figure \ref{PotentialOutcomes}. 

In the warm-start data, $x_{t,0}$ and $x_{t,1}$ are generated from a truncated normal distribution $\mathcal{N}(0, 1)$ on the interval $(-1.15, -0.85)$, while in subsequent data $x_{t,0}$ and $x_{t,1}$ are drawn from $\mathcal{N}(0, 1)$ without the truncation.
Each one of the 50 warm-start contexts is assigned to one of the three arms at random with equal probability. 
Note that the warm-start contexts belong to a region of the context space where the reward surfaces do not change much with the context. 
Therefore, when training the reward model for the first time, the estimated reward of arm $a = 2$ (blue) is the highest, the one of arm $a = 1$ (yellow) is the second highest and the one of arm $a = 0$ (red) is the lowest across the context space. 

We run our experiment with a learning horizon $T = 10000$. The regularization parameter $\lambda$, which is present in all algorithms, is chosen via cross-validation every time the model is updated. The constant $\alpha$, which is present in all algorithms, is optimized among values $0.25, 0.5, 1$ in the Thompson sampling bandits (the value $\alpha=1$ corresponds to standard Thompson sampling, \cite{chapelle-tsucb} suggest that smaller values may lower regret) and among values $1, 2, 4$ in the UCB bandits \cite{chapelle-tsucb}. The propensity threshold $\gamma$ for BLTS and BLUCB is optimized among the values $0.01, 0.05, 0.1, 0.2$.

\subsubsection{Well-Specified Outcome Models}\label{well}

\begin{figure}[t]	
	\centering
	\subfloat[Well-specified LinTS]{\includegraphics[width=0.95\columnwidth]{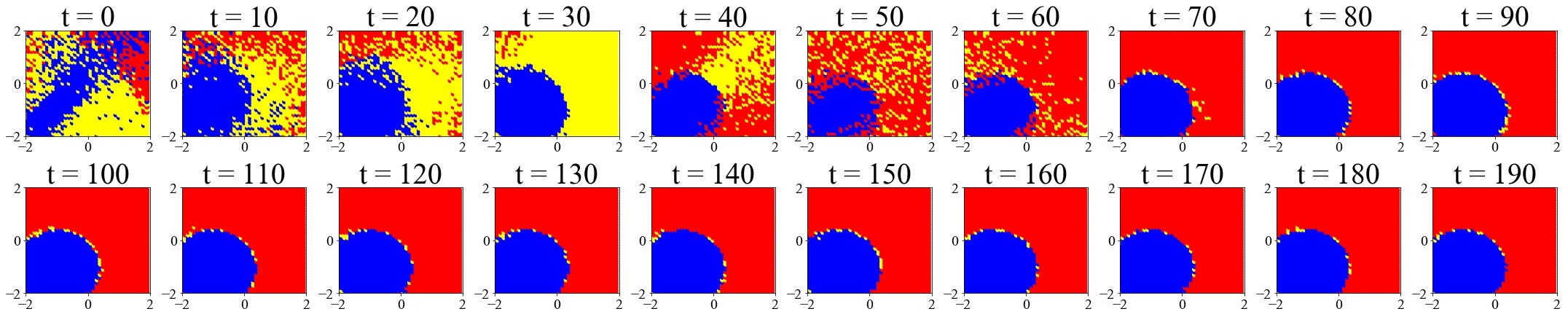}
		\label{RidgeTS-SecondOrder-PWFalse}} \\
	
	\subfloat[Well-specified LinUCB]{\includegraphics[width=0.95\columnwidth]{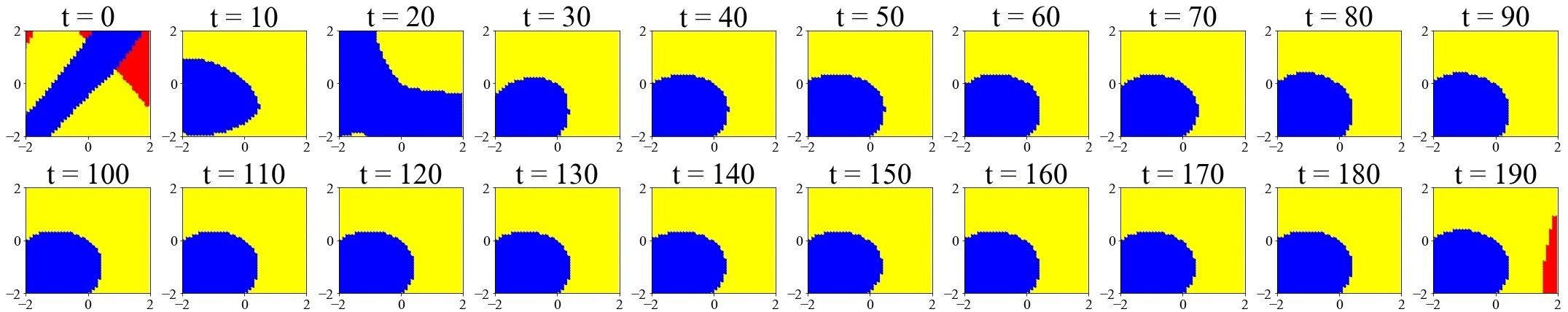}
		\label{RidgeUCB-SecondOrder-PWFalse}} \\
	\subfloat[Well-specified BLTS]{\includegraphics[width=0.95\columnwidth]{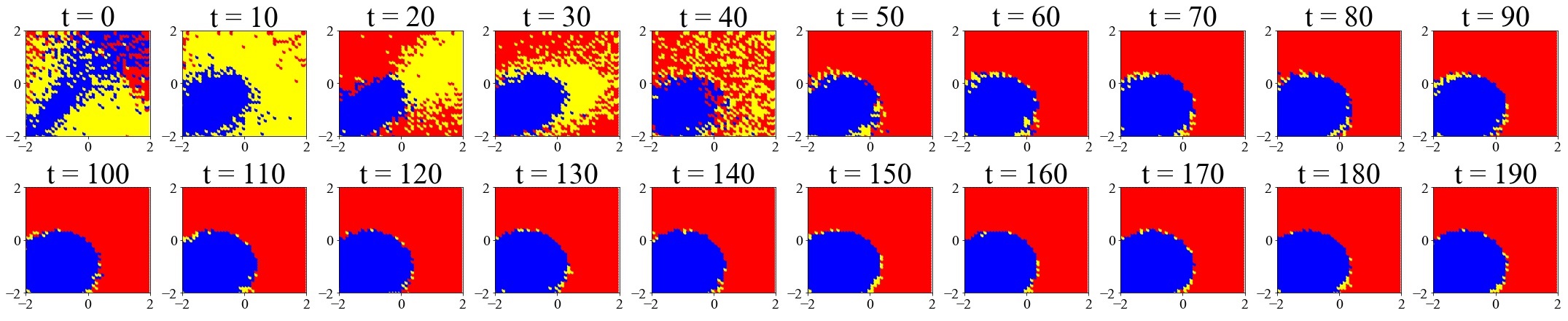}
		\label{RidgeTS-SecondOrder-PWTrue}}
	
	\subfloat[Well-specified BLUCB]{\includegraphics[width=0.95\columnwidth]{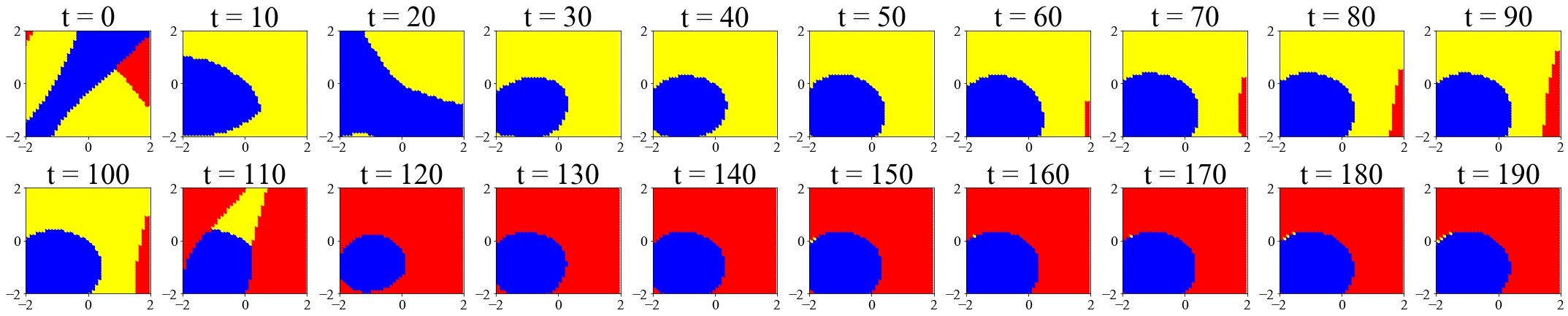}
		\label{RidgeUCB-SecondOrder-PWTrue}} \\
	\caption{Evolution of the arm assignment in the context space for well-specified LinTS, LinUCB, BLTS, BLUCB.}
\end{figure}

\begin{figure}[t]	
	\centering
	\subfloat[Mis-specified LinTS]{\includegraphics[width=0.95\columnwidth]{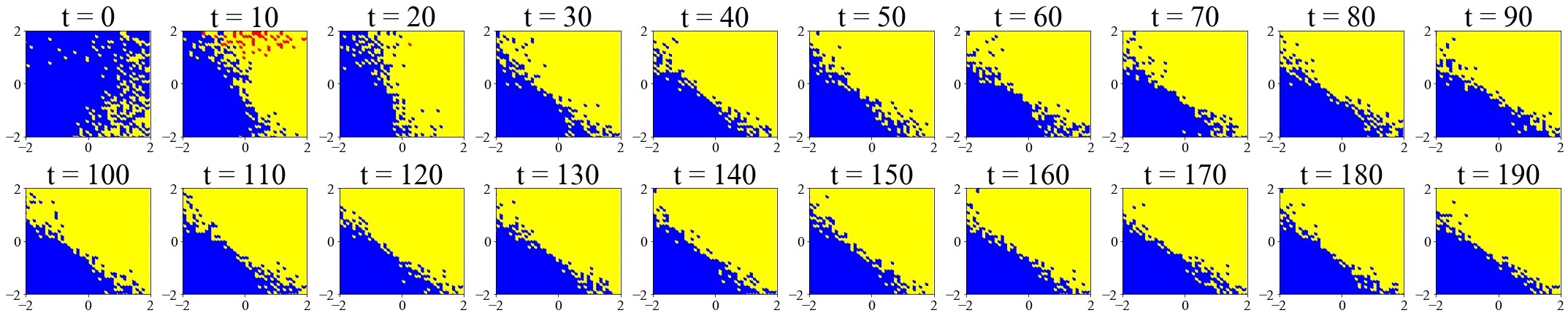}
		\label{RidgeTS-FirstOrder-PWFalse}} \\
	\subfloat[Mis-specified LinUCB]{\includegraphics[width=0.95\columnwidth]{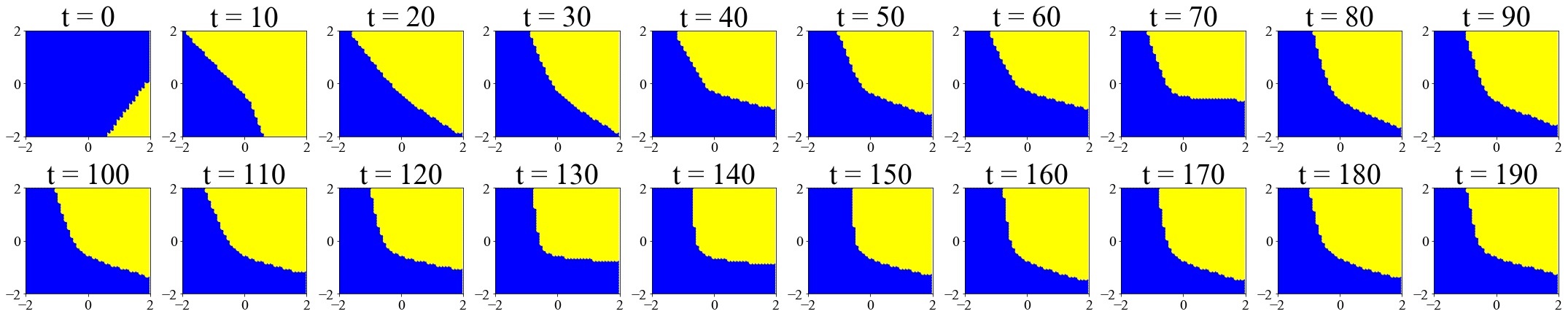}
		\label{RidgeUCB-FirstOrder-PWFalse}} \\
	\subfloat[Mis-specified BLTS]{\includegraphics[width=0.95\columnwidth]{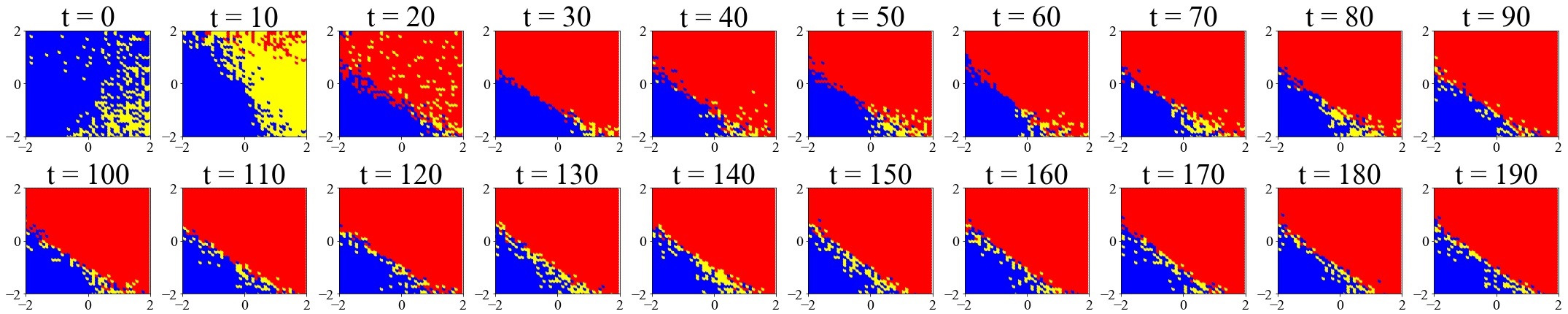}
		\label{RidgeTS-FirstOrder-PWTrue}} \\
	\subfloat[Mis-specified BLUCB]{\includegraphics[width=0.95\columnwidth]{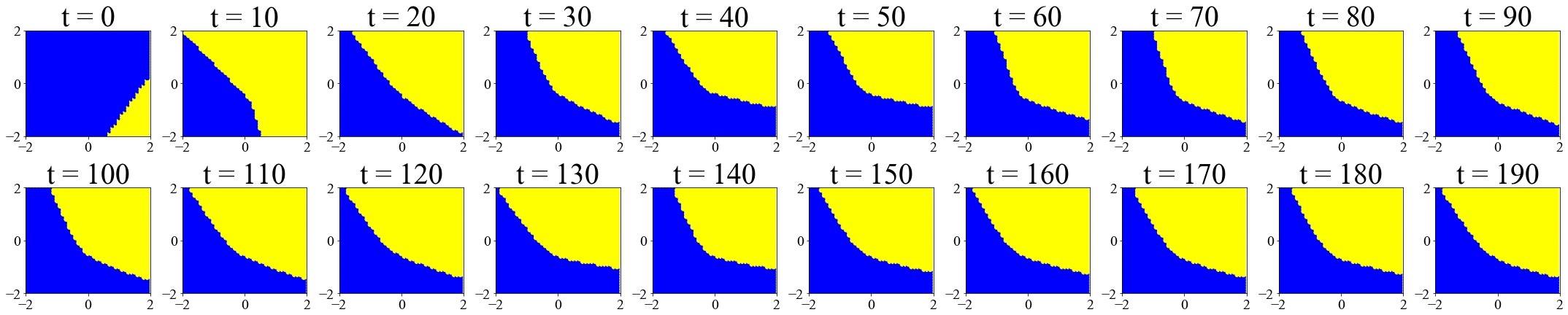}
		\label{RidgeUCB-FirstOrder-PWTrue}}
	\caption{Evolution of the arm assignment in the context space for misspecified LinTS, LinUCB, BLTS, BLUCB.}
\end{figure}

In this section, we compare the behavior of LinTS, LinUCB, BLTS and BLUCB when the outcome model of the contextual bandits is well-specified, i.e., it includes both linear and quadratic terms.
Note that this is still in the domain of linear contextual bandits, if we treat the quadratic terms as part of the context.

First, we compare LinTS and LinUCB. 
Figure \ref{RidgeTS-SecondOrder-PWFalse} shows that 
the uncertainty and the stochastic nature of LinTS leads to a ``dispersed'' assignment of arms $a=1$ and $a=2$ and to the crucial assignment of a few contexts to arm $a = 0$. 
This allows LinTS to start decreasing the bias in the estimation of all three arms.
Within the first few learning observations, LinTS estimates the outcome models of all three arms correctly and finds the optimal assignment. 
On the other hand, Figure \ref{RidgeUCB-SecondOrder-PWFalse}, shows that the deterministic nature of LinUCB assigns entire regions of the context space to the same arm. 
As a result not enough contexts are assigned to $a=0$ and LinUCB delays the correction of bias in the estimation of this arm.  
Another way to understand the problem is that the outcome model in the LinUCB bandit has biased coefficients combined with estimated uncertainty that is too low to incentivize the exploration of arm $a = 0$ initially. 
LinUCB finds the correct assignment after 240 observations.

Second, we study the performance of  BLTS and BLUCB.
In Figure \ref{RidgeUCB-SecondOrder-PWTrue}, we observe that balancing has a significant impact on the performance of UCB, since BLUCB finds the optimal assignment after 110 observations, much faster than LinUCB.
This is because the few observations of arm $a = 0$ outside of the context region of the warm-start batch are weighted more heavily by BLUCB. 
As a result, BLUCB, despite its deterministic nature which complicates estimation, is able to reduce its bias more quickly via balancing 
Figure \ref{RidgeTS-SecondOrder-PWTrue} shows that BLTS is also able to find the optimal assignment a few observations earlier than LinTS.

The first column of Table \ref{RidgeTSUCB-Percentages} shows the percentage of simulations in which LinTS, LinUCB, BLTS and BLUCB find the optimal assignment within $T = 10000$ contexts for the well-specified case.
BLTS outperforms all other algorithms by a large margin.

\subsubsection{Mis-Specified Outcome Models}

We now study the behavior of LinTS, LinUCB, BLTS and BLUCB when the outcome models include only linear terms of the context and therefore are misspecified.
In real-world domains, the true data generative process is complex and very difficult to capture by the simpler outcome models assumed by the learning algorithms. 
Hence, model mismatch is very likely. 

We first compare LinTS and LinUCB.
In Figures \ref{RidgeTS-FirstOrder-PWFalse}, \ref{RidgeUCB-FirstOrder-PWFalse}, we see that during the first time periods, both bandits assign most contexts to arm $a = 2$ and a few contexts to arm $a = 1$. 
LinTS finds faster than LinUCB the linearly approximated area in which arm $a = 2$ is suboptimal. 
However, both LinTS and LinUCB have trouble identifying that the optimal arm is $a = 0$. 
Due to the low estimate of $a = 0$ from the mis-representative warm-start observations, LinUCB does not assign contexts to arm $a = 0$ for a long time and therefore, delays to estimate the model of $a = 0$ correctly. 
LinTS does assign a few contexts to arm $a = 0$, but they are not enough to quickly correct the estimation bias of arm $a = 0$ either. 
On the other hand, BLTS is able to harness the advantages of the stochastic assignment rule of Thompson sampling. 
The few contexts assigned to arm $a = 0$ are weighted more heavily by BLTS. 
Therefore, as shown in Figure \ref{RidgeTS-FirstOrder-PWTrue}, BLTS corrects the estimation error of arm $a = 0$ and finds the (constrained) optimal assignment already after 20 observations. 
On the other hand, BLUCB does not handle better than LinUCB the estimation problem created by the deterministic nature of the assignment in the misspecified case, as shown in Figure \ref{RidgeUCB-FirstOrder-PWTrue}.
The second column of table \ref{RidgeTSUCB-Percentages} shows the percentage of simulations in which LinTS, LinUCB, BLTS and BLUCB find the optimal assignment within $T = 10000$ contexts for the misspecified case.
Again, BLTS has a strong advantage.

\subsubsection{}
This simple synthetic example allowed us to explain transparently where the benefits of balancing in linear bandits stem from. Balancing helps escape biases in the training data and can be more robust in the case of model misspecification. 
While, as we proved, balanced linear contextual bandits share the same strong theoretical guarantees, this indicates towards their better performance in practice compared to other contextual bandits with linear realizability assumption.
We investigate this further in the next section with an extensive evaluation on real classification datasets.

\begin{table}[h]
\centering
\begin{tabular}{l||c||c}
& Well-Specified & Mis-Specified \\
\hline 
LinTS & 84\% & 39\% \\ 
\hline
LinUCB & 51\% & 29\%  \\ 
\hline
{BLTS} & {92\%} & {58\%}  \\ 
\hline
BLUCB & 79\% & 30\% \\
\hline
\end{tabular}
\caption{Percentage of simulations in which LinTS, LinUCB, BLTS and BLUCB find the optimal assignment within learning horizon of $10000$ contexts}
\label{RidgeTSUCB-Percentages}
\end{table}

\subsection{Multiclass Classification with Bandit Feedback}\label{subsec:multi_class}

Adapting a classification task to a bandit problem is a common method for comparing contextual bandit algoriths \cite{dudik-offline-1}, \cite{agarwal-ilovetoconbandits}, \cite{bietti2018contextual}. In a classification task, we assume data are drawn IID from a fixed distribution: $(x, c) \sim D$, where $x \in \X$ is the context and $c \in {1, 2, \dots, K}$ is the class. The goal is to find a classifier $\pi: \X \rightarrow \{1, 2, \dots, K\}$ that minimizes the classification error $\E_{(x, c) \sim D}\mathbf{1}\left\{\pi(x) \neq c\right\}$. The classifier can be seen as an arm-selection policy and the classification error is the policy's expected regret. Further, if only the loss associated with the policy's chosen arm is revealed, this becomes a contextual bandit setting. So, at time $t$, context $x_t$ is sampled from the dataset, the contextual bandit selects arm $a_t \in \{1, 2, \dots, K\}$ and observes reward $r_t(a_t) = \mathbf{1}\left\{a_t = c_t\right\}$, where $c_t$ is the unknown, true class of $x_t$. 
The performance of a contextual bandit algorithm on a dataset with $n$ observations is measured with respect to the normalized cumulative regret, $\frac{1}{n} \sum_{t =1}^n{\left(1 - r_t(a_t)\right)}$.

We use 300 multiclass datasets from the Open Media Library (OpenML). The datasets vary in number of observations, number of classes and number of features. Table \ref{stats} summarizes the characteristics of these benchmark datasets. Each dataset is randomly shuffled. 

\begin{table}[H]
\centering
\begin{tabular}{|c|c|}
\hline
Observations & Datasets \\
\hline 
$\leq 100$ & 58 \\
\hline
$> 100$ and $\leq 1000$  & 152  \\ 
\hline
$> 1000$ and $\leq 10000$ & 57  \\ 
\hline
$> 10000$ & 33 \\
\hline
\end{tabular}
\\ \vspace{5pt}
\begin{tabular}{|c|c|}
\hline
Classes & Count \\
\hline 	
$2$ & 243 \\
\hline
$> 2 \text{ and } 10$  & 48  \\ 
\hline
$ > 10 $ & 9  \\
\hline
\end{tabular}
\hspace{5pt}
\begin{tabular}{|c|c|}
\hline
Features & Count \\
\hline 	
$\leq 10$ & 154 \\
\hline
$> 10 \text{ and } \leq 100$  & 106  \\ 
\hline
$> 100$  & 40  \\
\hline
\end{tabular}
\caption{Characteristics of the 300 datasets used for the experiments of multiclass classification with bandit feedback.}
\label{stats}
\end{table}

We evaluate LinTS, BLTS, LinUCB and BLUCB on these 300 benchmark datasets. We run each contextual bandit on every dataset for different choices of input parameters. The regularization parameter $\lambda$, which is present in all algorithms, is chosen via cross-validation every time the model is updated. The constant $\alpha$, which is present in all algorithms, is optimized among values $0.25, 0.5, 1$ in the Thompson sampling bandits \cite{chapelle-tsucb} and among values $1, 2, 4$ in the UCB bandits \cite{chapelle-tsucb}. The propensity threshold $\gamma$ for BLTS and BLUCB is optimized among the values $0.01, 0.05, 0.1, 0.2$.
Apart from baselines that belong in the family of contextual bandits with linear realizability assumption and have strong theoretical guarantees, we also evaluate the policy-based ILOVETOCONBANDITS (ILTCB) from \cite{agarwal-ilovetoconbandits} that does not estimate a model, but instead it assumes access to an oracle for solving fully supervised cost-sensitive classification problems and achieves the statistically optimal regret.

\begin{figure}[!htb]
\centering
\includegraphics[width=0.95\columnwidth]{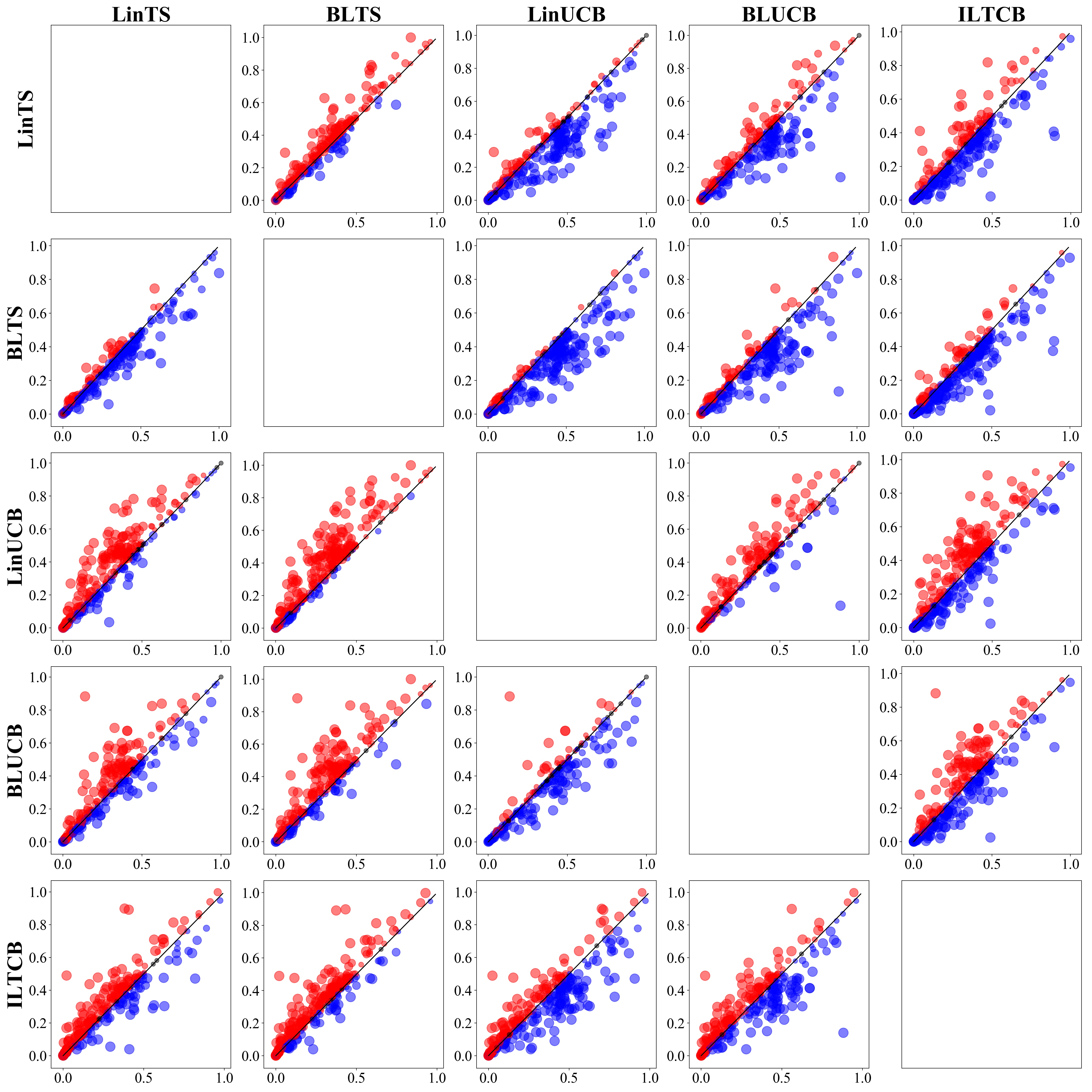}
\caption{Comparing LinTS, BLTS, LinUCB, BLUCB, ILTCB on 300 datasets. BLUCB outperforms LinUCB. BLTS outperforms LinTS, LinUCB, BLUCB, ILTCB.}
\label{fig:all_vs_all}
\end{figure}

Figure \ref{fig:all_vs_all} shows the pairwise comparison of LinTS, BLTS, LinUCB, BLUCB and ILTCB on the 300 classification datasets. Each point corresponds to a dataset. The $x$ coordinate is the normalized cumulative regret of the column bandit and the $y$ coordinate is the normalized cumulative regret of the row bandit. The point is blue when the row bandit has smaller normalized cumulative regret and wins over the column bandit. The point is red when the row bandit loses from the column bandit. The point's size grows with the significance of the win or loss. 

The first important observation is that the improved model estimation achieved via balancing leads to better practical performance across a large number of contextual bandit instances. Specifically, BLTS outperforms LinTS and BLUCB outperforms LinUCB.
The second important observation is that deterministic assignment rule bandits are at a disadvantage compared to randomized assignment rule bandits. The improvement in estimation via balancing is not enough to outweigh the fact that estimation is more difficult when the assignment is deterministic and BLUCB is outperformed by LinTS.
Overall, BLTS which has both balancing and a randomized assignment rule, outperforms all other linear contextual bandits with strong theoretical guarantees.
BLTS also outperforms the model-agnostic ILTCB algorithm.
We refer the reader to Appendix B of the supplemental material of the extended version of this paper \cite{dimakopoulou2017estimation} for details on the datasets.

\section{Closing Remarks}
Contextual bandits are poised to play an important role in a wide range of applications:
content recommendation in web-services, where the learner wants to personalize recommendations (arm) to the profile of a user (context) to maximize engagement (reward);  online education platforms, where the learner wants to select a teaching method (arm) based on the characteristics of a student (context) in order to maximize the student's scores (reward); and survey experiments, where the learner wants to learn what information or persuasion (arm) influences the responses (reward) of subjects as a function of their demographics, political beliefs, or other characteristics (context).
In these settings, there are many potential sources of bias in estimation of outcome models, not only due to the inherent adaptive data collection, but also due to mismatch between the true data generating process and the outcome model assumptions, and due to prejudice in the training data in form of under-representation or over-representation of certain regions of the context space.
To reduce bias, we have proposed new contextual bandit algorithms, BLTS and BLUCB, which build on linear contextual bandits LinTS and LinUCB respectively and improve them with balancing methods from the causal inference literature. 

We derived the first regret bound analysis for linear contextual bandits with balancing and we showed linear contextual bandits with balancing match the theoretical guarantees of the linear contextual bandits with direct model estimation; namely that BLTS matches the regret bound of LinTS and BLUCB matches the regret bound of LinUCB.
A synthetic example simulating covariate shift and model misspecification and a large-scale experiment with real multiclass classification datasets demonstrated the effectiveness of balancing in contextual bandits, particularly when coupled with Thompson sampling.

\section{Acknowledgments}
The authors would like to thank Emma Brunskill for valuable comments on the paper and John Langford, Miroslav Dud{\'\i}k, Akshay Krishnamurthy and Chicheng Zhang for useful discussions regarding the evaluation on classification datasets. 
This research is generously supported by ONR grant N00014-17-1-2131, by the Sloan Foundation, by the ``Arvanitidis in Memory of William K. Linvill'' Stanford Graduate Fellowship and by the Onassis Foundation.

\fontsize{9.5pt}{10.5pt} \selectfont
\bibliography{reference}
\bibliographystyle{aaai}
\end{document}